\documentclass[letterpaper, 10 pt, conference]{ieeeconf}  

\IEEEoverridecommandlockouts   

\usepackage{graphics} 
\usepackage{epsfig} 
\usepackage{amsmath} 
\usepackage{amssymb}  
\usepackage{gensymb}
\usepackage{multirow}
\usepackage{siunitx}
\usepackage{cite}
\usepackage{url}
\usepackage{caption}
\usepackage{titlesec}
\usepackage{placeins} 
\usepackage{soul}
\usepackage{float}

\usepackage{color} 

\title{\LARGE \bf
Multilaminate piezoelectric polymer actuators for insect-scale softrobots
}

\author{Nicholas Gunter$^{1}$, Heiko Kabutz$^{1,2}$, Alexander Hedrick$^{1}$, Kaushik Jayaram$^{1,2,*}$
\thanks{Any opinions, findings, and conclusions or recommendations expressed in this material are those of the authors(s) and do not necessarily reflect the views of the any funding agency. This work is partially funded through grants from US Army research office (ARO) Grant \# W911NF-23-1-0039 and NSF CAREER Award \# 2443869.}
\thanks{$^{1}$Animal Inspired Movement and Robotics Laboratory, Mechanical Engineering, University of Colorado Boulder, USA; $^{2}$Insect Inspired Intelligence Lab (III), Bioengineering, Imperial College London, UK} 
\thanks{$^{*}${For correspondence, \tt\footnotesize k.jayaram@imperial.ac.uk}}%
}
\begin{document}

\maketitle
\thispagestyle{empty}
\pagestyle{empty}

\begin{abstract}
Piezoelectric ceramics enable fast, precise, and powerful actuation with integrated sensing in insect-scale robots. However, they are brittle, failing at low strains limiting its applications involving high impacts, unexpected loading, or soft robotics. To address this limitation, we investigate polyvinylidene fluoride (PVDF) as a compliant alternative: it tolerates substantially larger strains, but its low piezoelectric coefficient limits the output of conventional single-layer actuators. We demonstrate a novel manufacturing process for multilayer PVDF actuators that utilizes off the shelf materials and provides parallel bussing of signal voltage to reduce integration complexity. Using first principles models validated with experiments, we show a tradeoff between force (linear or superlinear increase with layer count) and displacement (cubic decrease with layer count) for the actuator geometry chosen. Our work positions PVDF actuation an optimization problem for future robotic applications with clear design knobs available to meet performance demands. Further, we show that using thinner PVDF layers within a stack increases the specific power density of the actuators making them ideal for size, weight, and power (SWAP) limitations in small scale robotics. Among the actuators fabricated, $>$9 mm of free deflection, $>$20 mN of blocked force, and $>$500 Hz bandwidth were achieved. To illustrate their potential in future robust robotic platforms, we integrate our actuators into a low-profile insect-scale robot that leverages resonance to achieve planar locomotion while being robust to large external impacts.
\end{abstract}

\vspace{-1mm}

\section{Introduction}

Piezoelectric actuators dominate actuation at insect scales because their intrinsic electromechanical coupling delivers precise, rapid, controllable motion with integrated sensing \cite{kabutz_integrated_2025}. For robots approaching insect size, lead zirconate titanate (PZT) is the primary choice for both flight \cite{jafferis_untethered_2019, chukewad_robofly_2021} and walking \cite{doshi_effective_2019, kabutz_design_2023}, offering high piezoelectric coefficients, large electromechanical coupling factors, resulting in device operational bandwidths \cite{Snis2008_MonolithicPVDFTrFE}. 

\begin{figure} [!tb]
	\centering

    \includegraphics[width=0.9\linewidth]{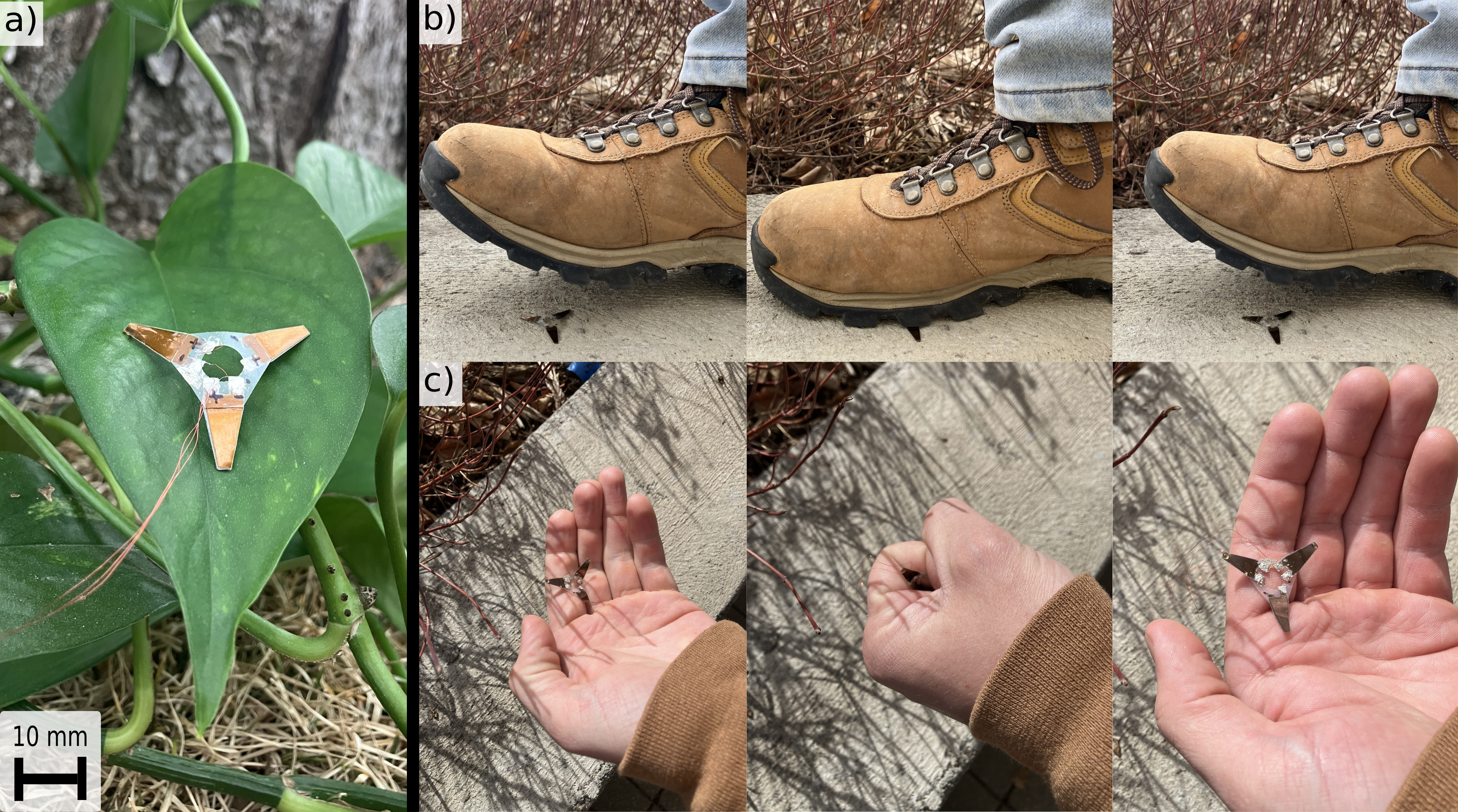}
    \caption{a) PVDF Tripod crawling on Pothos plant. b) PVDF Tripod being stepped on. c) PVDF Tripod being crushed in hand.}
    \label{fig:RobotOverview}
\end{figure}
However, PZT is brittle (maximum allowable strains \SI{0.3}{\percent} \cite{PANDA2023100226}) for robots that must survive repeated collisions during high-speed running \cite{goldberg_gait_2017} and operate under dynamic loading in cluttered environments \cite{kabutz_mclari_2023}. In addition, PZT is dense (\SI{7.6}{\gram/\centi\meter^3}) for insect-scale systems that aim to maximize the propulsion (or payload) to weight ratio, leading researchers to explore lightweight and compliant alternatives \cite{chen_controlled_2019}. 
Recent soft electroactive systems, including shape memory polymers (SMPs, \cite{lendlein2018reprogrammable}), dielectric elastomers (DEAs, \cite{ren_highlift_2022}) and hydraulically amplified electrostatic actuators (HASELs, \cite{kellaris_peano-hasel_2018, acome_hydraulically_2018}), each have a demonstrated niche for which they are uniquely suited \cite{chen2026call}. In particular, DEAs and HASELs can achieve strains exceeding \SI{100}{\percent} and have been recorded among the highest specific power densities of any soft actuation technology~\cite{Costa2023_ChemRev_PVDF_EAPs}\cite{KIM202219}. Despite growing promise, soft actuators currently are have distinct disadvantages for small scale field robotics. For example, HASELs have high voltage requirements (in \SI{}{\kilo\volt}s) at the macroscale, and greatly reduced power density at milliscale. Additionally, their force generation relies on the dynamic pressure of fluid in a pouch which can be torn or punctured \cite{KIM202219}. Although the voltage requirements of DEA actuators are decreasing (\SI{0.3}{}-\SI{2}{\kilo\volt}, \cite{peng2026lowvoltage}) and they have greater power density than similar sized HASELs, they have a catastrophic mode of failure known as "pull in" where the electrostatic force applied to the polymer is greater than the restorative force of the elastomer \cite{PLANTE20067727}. These actuators typically suffer from poor manufacturability when scaled down resulting in lower performance and increase failure \cite{shmulevich2026microfabrication}. 
These tradeoffs indicate a sparsely populated actuator design space that is ripe for exploration: moderate force, high bandwidth, low voltage drive, and high mechanical strain tolerance sufficient to survive impact. 

\begin{figure*} [tb]
    \centering
    \includegraphics[width=0.8\textwidth]{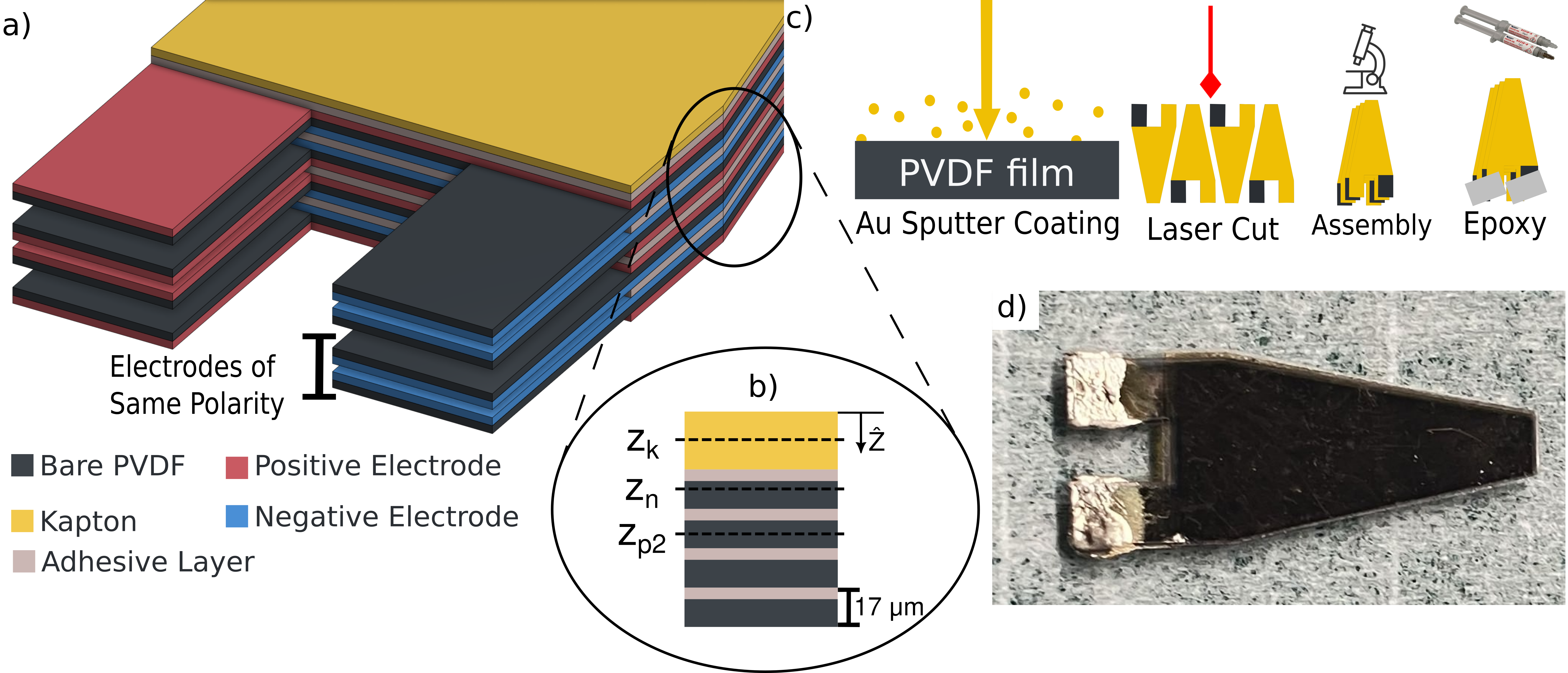}
    \caption{Multilayer actuator design and manufacturing. a) \textbf{Asymmetric stacking} enables parallel bussing of voltage. b) Stacking representation used to develop theory (Section \ref{sec: Theory}). c) Manufacturing steps. d) Completed prototype.}
    \label{fig:Manufacturing}
\end{figure*}

We contend that PVDF and its copolymers exhibit a range of attractive properties that enable them to potentially bridge the performance gaps described above \cite{Punetha2026}. 
PVDF sustains strains over \SI{2}{\percent} (up to \SI{5}{\percent}) \cite{ahbab2025pvdf} for repeated large-deflection bending without fracture, permits sustained resonant operation ($Q \approx 8-15$, \cite{Fiorillo1992_PVDFRangeSensor}) without self-damage, withstands fields exceeding \SI{100}{\kilo\volt/\milli\meter}, has low density (\SI{1.8}{\gram/\centi\meter^3}), and can serve simultaneously as actuator and sensor due to inherent electromechanical coupling \cite{kabutz_integrated_2025}. 

Despite these advantages, the relatively low piezoelectric coefficient of commercially available PVDF ($\sim$\SI{25}{\pico\meter/\volt}) yields insufficient force and displacement for useful robotic work in single-layer configurations. 
To remedy this, researchers have resorted to stacking multiple active PVDF layers driven in parallel electrically \cite{Snis2008_MonolithicPVDFTrFE}. Recently, Toinet et al. \cite{toinet2024design, toinet2025ultra} demonstrated screen-printed multi-stack P(VDF-TrFE) and terpolymer actuators optimized for steerable surgical guidewires.
Although successful, these approaches need to be adapted for microrobotics. Complex fabrication processes involving successive deposition of active polymers and electrodes at \SI{150}{\celsius} under vacuum followed by post-lamination poling require specialized equipment under clean room conditions, potentially limiting widespread adoption.
Additionally, the low force output (not explicitly characterized in \cite{toinet2024design}) and  bandwidth (below \SI{100}{\hertz}) consistent with their use case is not ideal for resonant operations of high-speed flapping or legged robots. In fact, \cite{ZhuTakeoff2026} has directly called for further research into off the shelf multilayered, soft piezoelectric actuators.

In this work, we target the multilayer architecture to increase PVDF actuator performance using an assembly-route fabrication process compatible with laminate microrobots. In Section \ref{sec: Manufacturing}, we adapt the femtosecond micromachining \cite{hedrick2024femtosecond} and laminate stacking \cite{jafferis_streamlined_2021} fabrication previously developed for multilayer PZT actuators to create multilaminate stacks of PVDF unimorph beam actuators using off-the-shelf polarized thin films. The design allows for parallel application of voltage across each layer for ready integration into existing robotic designs \cite{kabutz_mclari_2023}. Our choice of using commercially pre-poled, uniaxially stretched PVDF allows easy fabrication outside cleanroom environments. To prevent depoling near its Curie point (and degradation beyond \SI{80}{\celsius}, \cite{kabutz_integrated_2025}), deposition-and-anneal methods cannot be used. We instead laminate pre-poled films entirely at room temperature using thin PET-based pressure adhesive interlayers, with gold electrodes sputtered through asymmetric shadow masks and an uncoated perimeter buffer zone that eliminates interlayer shorting. 
We fabricate 4- and 8-layer devices and demonstrate that their blocked force and free displacement performance are predictable from first-principles models derived from solid mechanics in Section \ref{sec: Theory}. This multilaminate architecture introduces two tunable design parameters, the number of layers and layer thickness, which enable systematic customization of actuator performance for future small-scale robotic applications.
Other contributions of this work include the detailed displacement and force characterization of PVDF laminate actuators (Section \ref{sec: Experiment}) as a function of geometry (layer thickness, layer number) and operational dynamics (voltage, frequency).
We show that the blocked force increases approximately linearly with the number of active layers when driven in parallel, reducing the overall voltage required to achieve a given output force relative to the use of bulk materials. This multilaminate design strategy enhances the force density and scalability comparable to other best-in-class miniature actuators while maintaining relatively low driving voltage requirements, making PVDF actuators particularly compelling for microrobotic applications where stroke, robustness, and ease of integration outweigh the need for extreme force output. We further demonstrate that our PVDF actuators achieve {over \SI{9}{\milli\meter} free deflection} at resonance, {\SI{20}{\milli\newton} of blocked force} at resonance, {driving voltages as low as \SI{150}{\volt}}, {and overall bandwidths exceeding \SI{500}{\hertz}} depending on layer count and thickness. Finally, to illustrate their potential for robotic integration, we present a lightweight (\SI{117}{\milli\gram}) planar translation microrobot driven by resonant PVDF actuation (Section \ref{sec: TriBot}) that remains robust even after significant impact loading.

\section{Actuator Design and Fabrication}
\label{sec: Manufacturing}
One way to address the relatively weak electromechanical coupling of PVDF is to increase the amount of active piezoelectric material in an actuator \cite{toinet2024design}. Thus, a multilaminate design where parallel driving voltage is applied to each layer separately was devised as described below. The final prototype actuator parameters are listed in Table \ref{tab:parameters}.
\begin{table}[!ht]
    \centering
    \caption{Actuator parameters (-- beam dependent values).}
    \begin{tabular}{|c|c|c|c|}
        \hline
        Parameter & Symbol & Value & Units \\
        \hline
        \multicolumn{4}{|c|}{PVDF Material Properties} \\
        \hline
        Curie Temperature & $T_c$ & 80 & \celsius \\
        Piezoelectric Coupling Coefficient & $d_{31}$ & 25 & pC/N \\
        Young's Modulus & $E_p$ & 2.8 & GPa \\
        Layer Thickness & $t_{ell}$ & $\in \{7, 12\}$ & $\mu$m \\
        Number of Layers & $N$ & $\in\{4, 8\}$ & unitless\\
        \hline
        \multicolumn{4}{|c|}{Kapton Material Properties} \\
        \hline
        Young's Modulus & $E_k$ & 4 & Gpa \\
        Thickness & $t_k$ & 25 & $\mu$m \\
        Centroid Position & $z_k$ & 12.5 & $\mu$m \\
        \hline
        \multicolumn{4}{|c|}{PET Adhesive Material Properties} \\
        \hline
        Young's Modulus & $E_a$ & 3 & GPa \\
        Thickness & $t_a$ & 5 & $\mu$m \\ 
        \hline
        \multicolumn{4}{|c|}{Beam Parameters} \\
        \hline
        Actuator Length & $L$ & 10 & mm \\
        Flexural Stiffness & $EI_{eq}$ & ---- & Nm$^2$\\
        Neutral Axis Position & $z_n$ & ---- & m\\
        \hline    
    \end{tabular}
    \label{tab:parameters}
\end{table}

\subsection{Multilayer Asymmetric Stacking Architecture}
Our actuator (Fig. \ref{fig:Manufacturing}) consists of a passive stiffening layer, electrode coated PVDF layers and stacked with adhesive interlayers. We design the actuator in a cantilever beam form factor to maximize displacement at resonance similar to previous studies \cite{wood2005optimal}.

\textit{Flexible design with improved manufacturability.} We used commercially available (for consistent electromechanical properties across batches) uniaxially oriented PVDF films (PolyK Technologies) with thicknesses $t_\ell \in \{7,\,12\}\,\mu$m. 
Film thickness is a key design parameter for scaling. Thinner films experience a higher electric field for the same applied voltage, according to $E = V / t_\ell$, resulting in a proportionally higher actuation strain. In contrast, thicker films provide greater flexural stiffness but lower field strength for a given voltage. This tradeoff is reflected in the experimental results presented in Sec. \ref{sec: Experiment}.

\textit{Mitigating dielectric breakdown.}
 Electrical shorting is the primary limitation in thin-film multilaminate architectures, where the breakdown voltages of air and adhesive are not sufficient to prevent dielectric failure on their own. To address this, we implemented an \textit{asymmetric stacking architecture} in which each electrode is neighboured by one of the same polarity (Fig. \ref{fig:Manufacturing}a. This configuration allows for positive electrode routing to one side and negative routing to the other. The layers in each tab are collectively bonded using silver epoxy. This arrangement eliminates successive interlayer shorting while maintaining a simple and scalable electrical interface. Electrical shorting around the edges of the actuator was eliminated with a laser etched buffer zone (15 $\mu$m) without electrodes along the perimeter.

\subsection{Repeatable and Scalable Manufacturing}

Multilayer actuators were manufactured by laminating individually patterned PVDF films with a thin adhesive layer at room temperature (Fig. \ref{fig:Manufacturing}c). We developed a fabrication process to minimize thermal depoling which is a major problem limiting PVDF manufacturing. When exposed to temperatures near or above its Curie point (\SI{80}{\celsius}) which are common with conventional lamination or laser machining processes \cite{kabutz_integrated_2025}, PVDF's piezoelectric performance rapidly degrades. To address this,  
the multilayer stacks were fabricated entirely at room temperature using a thin-film, PET-based double-sided tape (Nitto Denko) to bond the layers to each other. This approach preserved the polarization state of the PVDF films while ensuring consistent interfacial adhesion and layer alignment. Additionally, to create custom electrodes on the surface of PVDF, we used sputter coating through shadow masks, where bare PVDF films were cut to size with the laser cutter, masked and sputtered with gold. This approach improves manufacturing repeatability and overall success relative to our previous process of lasermachining pre-metallized films \cite{kabutz_integrated_2025} which risked local heating and electromechanical failure from imperfect ablation.

A 25 $\mu$m polyimide ({Dupont Inc.}) sheet was used as the passive stiffening stiffening layer due to its moderate stiffness and good adhesion to the Nitto adhesive. 

Finally, the leads of the alternating electrode architecture were bonded with silver epoxy to allow all PVDF layers to experience the full electric field. This configuration ensures that each layer contributes maximally to the work output while simplifying the electrical connection for ease of use in robotic platforms.

\section{Multilayer Unimorph Composite Beam Model}
\label{sec: Theory}

Expanding on classical models of cantilever actuators \cite{toinet2024design, jafferis_design_2015}, we develop a custom composite Euler-Bernoulli beam model to be used as a tool for actuator design. 

\newcommand{\EIeq}{EI_{\text{eq}}}
A unimorph PVDF actuator can be modelled as a composite Euler–Bernoulli beam with the multilayer (adhesive + active  with coupling coefficient, $d_{31}$) stack on top of the stiffening layer. 

The beam (total thickness $t_\ell$) is treated as a stack of $1+2N$ discrete layers (Fig. \ref{fig:Manufacturing}b): the Kapton substrate, and $N$ adhesive (\textit{a}) or active (\textit{p}) layer pairs for $j = 1,\dots,N$. Measuring $z$ from the bottom of the substrate, the centroids of the $j$-th adhesive ($z_{a,j}$) and active layer ($z_{p,j}$) are, respectively,
\begin{equation}
\begin{split}
z_{a,j} &= t_k + (j-1)(t_a + t_\ell) + \frac{t_a}{2}, \\
z_{p,j} &= t_k + j\,t_a + (j-1)\,t_\ell + \frac{t_\ell}{2}
\end{split}
\label{eq:layer_z}
\end{equation}
Adhesive and actives layer with width ($w$) have a  cross-sectional area $A_a = w\,t_a$ and $A_p = w\,t_\ell$, respectively. The neutral axis ($z_n$) and equivalent flexural stiffness ($EI_{\text{eq}}$) are:
\begin{equation}
z_n = \frac{E_k A_k z_k + \sum_{j=1}^{N} E_a A_a z_{a,j} + \sum_{j=1}^{N} E_p A_p z_{p,j}}{E_k A_k + N E_a A_a + N E_p A_p}
\label{eq:NA}
\end{equation}
\begin{equation}
\begin{split}
EI_{\text{eq}} = {}& E_k I_k + E_k A_k (z_k - z_n)^2 \\
& + \sum_{j=1}^{N}\Big[E_a I_a + E_a A_a (z_{a,j}-z_n)^2\Big] \\
& + \sum_{j=1}^{N}\Big[E_p I_p + E_p A_p (z_{p,j}-z_n)^2\Big]
\end{split}
\label{eq:EIeq}
\end{equation}
where $I_k$, $I_a$, and $I_p$ are the second moments of area of the stiffener, adhesive, and active layer about their own centroids.

\begin{align}
M_{\text{act}} &= \epsilon^{\ast}\, E_p A_p \sum_{j=1}^{N} (z_{p,j} - z_n) \notag \\
&= d_{31}\,\frac{V}{t_\ell} E_p A_p \sum_{j=1}^{N} (z_{p,j} - z_n)
\label{eq:M_act}
\end{align}

Therefore, the blocked force ($F_{\text{b}}$) and free displacement ($\delta_{\text{free}}$) of the actuator are:

\begin{subequations}
\begin{align}
\delta_{\text{free}} &= \frac{L^2}{2} \, \frac{M_{\text{act}}}{EI_{\text{eq}}} = \frac{L^2\,d_{31}\,V\,E_p\,A_p}{2\,t_\ell\,EI_{\text{eq}}} \sum_{j=1}^{N} (z_{p,j} - z_n) \label{eq:free displacement combined} \\
F_{\text{b}} &= \frac{3\, M_{\text{act}}}{2L} = \frac{3\,d_{31}\,V\,E_p\,A_p}{2\,t_\ell\,L} \sum_{j=1}^{N} (z_{p,j} - z_n) \label{eq:blocked force combined}
\end{align}
\end{subequations}

\textit{Quasistatic Force--Displacement Tradeoff.}
Eqns. \ref{eq:free displacement combined} and \ref{eq:blocked force combined} capture the primary electromechanical coupling behaviour and are experimentally validated in Sec. \ref{sec: Experiment}. They
indicate that {higher voltages}, {larger piezoelectric coefficients}, and {thinner layers} improve overall free displacement and blocked force. They also indicate that thicker beams (i.e., more layers) displace lesser but produce higher forces.
This tradeoff between force generation and displacement amplitude defines an actuator’s optimal geometry for a given application.

\textit{Resonance.}
We estimate the resonance frequency of a cantilever unimorph with mass per unit length, $\rho_\text{eq}A$, as:
\begin{equation}
    f_\text{theory} \approx \frac{1.875^2}{2\pi L^2}\sqrt{\frac{(EI)_\text{eq}}{\rho_\text{eq}A}}
    \label{eq:freq}
\end{equation}
We experimentally quantify the quality factor $Q$, which governs the trade-off between improved resonant performance and bandwidth. We expect displacement and force amplification proportional to Q at resonance \cite{ahbab2025pvdf}.

\textit{Power Density Characterisation.}
We characterise the actuator's mechanical power output in two ways: a load-line estimate of the maximum (peak) power from blocked force and free displacement\cite{piezosystems_actuators}, and a measured continuous (cycle-averaged) power from loaded cyclic operation.

\noindent\textit{(1) Load Line Method.} For quasi-linear actuators, the force--displacement relationship is a straight line between the blocked force $F_b$ (at zero displacement) and the free displacement $\delta_{free}$ (at zero force). The mechanical work per cycle is maximised when the actuator delivers half the blocked force over half the free displacement, yielding a peak output power $P_{peak}$ at drive frequency $f$ as:
\begin{equation}
    P_{peak}= \frac{F_b\, \delta_{free}\, f}{4\, m}
    \label{eq: peak power}
\end{equation}
We use this metric to compare with other state-of-the-art (SoA), soft milli-actuators in Table \ref{tab:actuator_comparison}.

\noindent\textit{(2) Continuous Power.} 
We compute the cycle-averaged mechanical power during sustained operation as:
\begin{equation}
    P = \frac{\int_{t_0}^{t_f} \vec{F}\cdot \vec{v}\, dt}{t_f-t_0}
    \label{eq: power}
\end{equation}
where $\vec{F}$ is the instantaneous tip force, $\vec{v}$ is the instantaneous tip velocity, and $t_0$ and $t_f$ are the initial and final times of the actuation cycle, respectively.
Each actuator was loaded with a tip mass (10--15\,mg) to half its displacement, corresponding to the matched-load condition of Eq.~\ref{eq: peak power}. This metric better reflects typical operation and is used to compare actuator samples within this work.

\section{Actuator Performance}
\label{sec: Experiment}

This section evaluates the performance of multilaminate piezoelectric polymer actuators. Using a 4-\SI{12}{\micro\meter}-layer (4 layers, \SI{12}{\micro\meter} each) example, we analyse quasi-static and dynamic actuation characteristics in Sec. \ref{sec:blocked_force}. We then extend the analysis to actuators with other layer counts and thicknesses in Sec. \ref{sec:layers}.

\subsection{Experimental Setup}
Tip displacement was measured using a time-of-flight (ToF) laser displacement sensor (Keyence LK-HD500), while blocked force was recorded using a miniature force sensor (LSB200, Futek). A high-speed camera (Phantom V720) was used concurrently to verify displacement measurements and confirm minimal out-of-plane motion. All data were acquired in real time using a Speedgoat real-time computer.

\begin{figure}[!htb]
	\centering
	\includegraphics[width=0.7\linewidth]{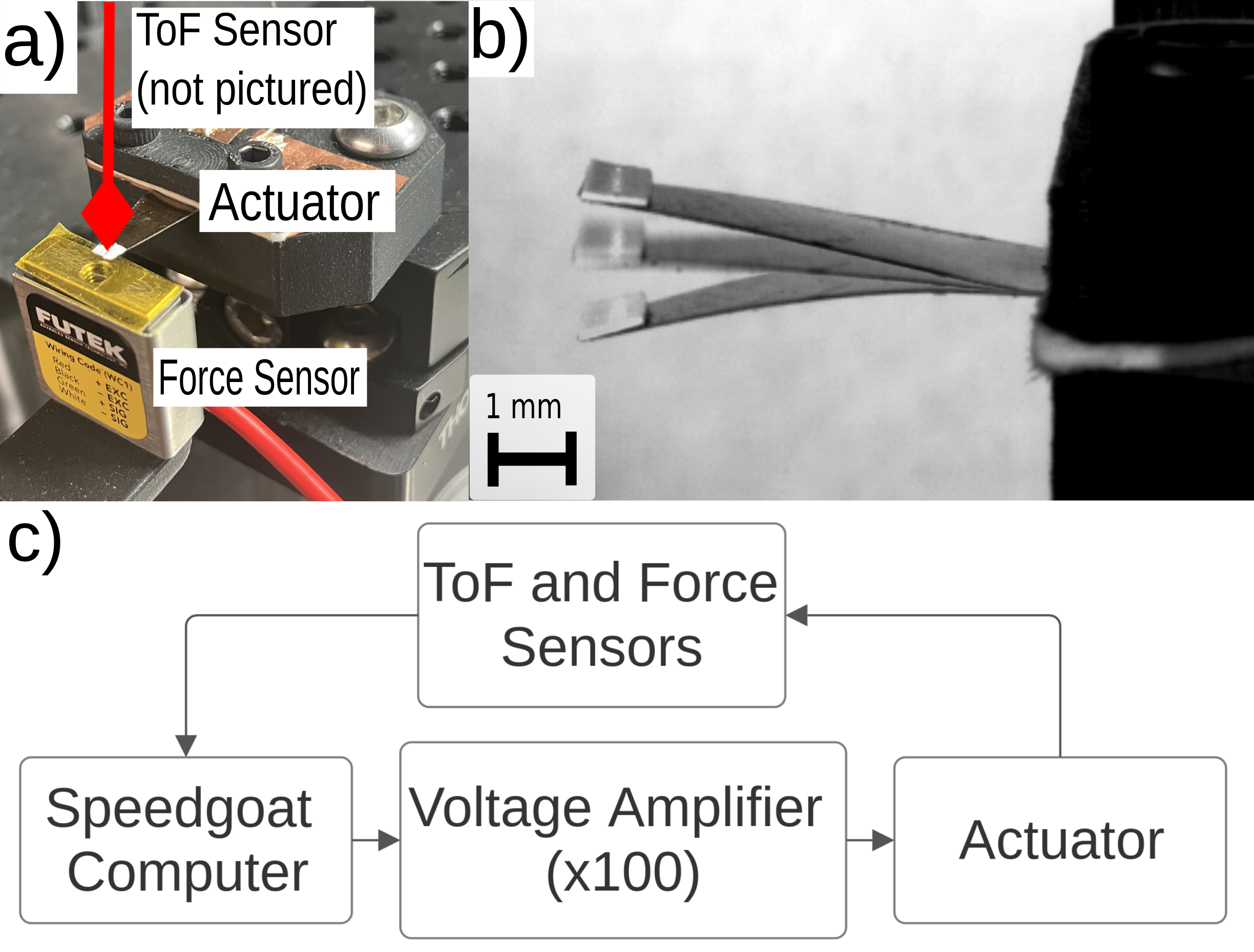}
	\caption{a) Experimental charaterization setup. b) Free displacement example. c) Drive configuration block diagram.}
	\label{fig:ExperimentalSetup}
    \vspace{-0.5cm}
\end{figure}

\subsection{Blocked Force and Free Displacement}
\label{sec:blocked_force}

\begin{figure}[!b]
	\centering
	\includegraphics[width=0.9\linewidth]{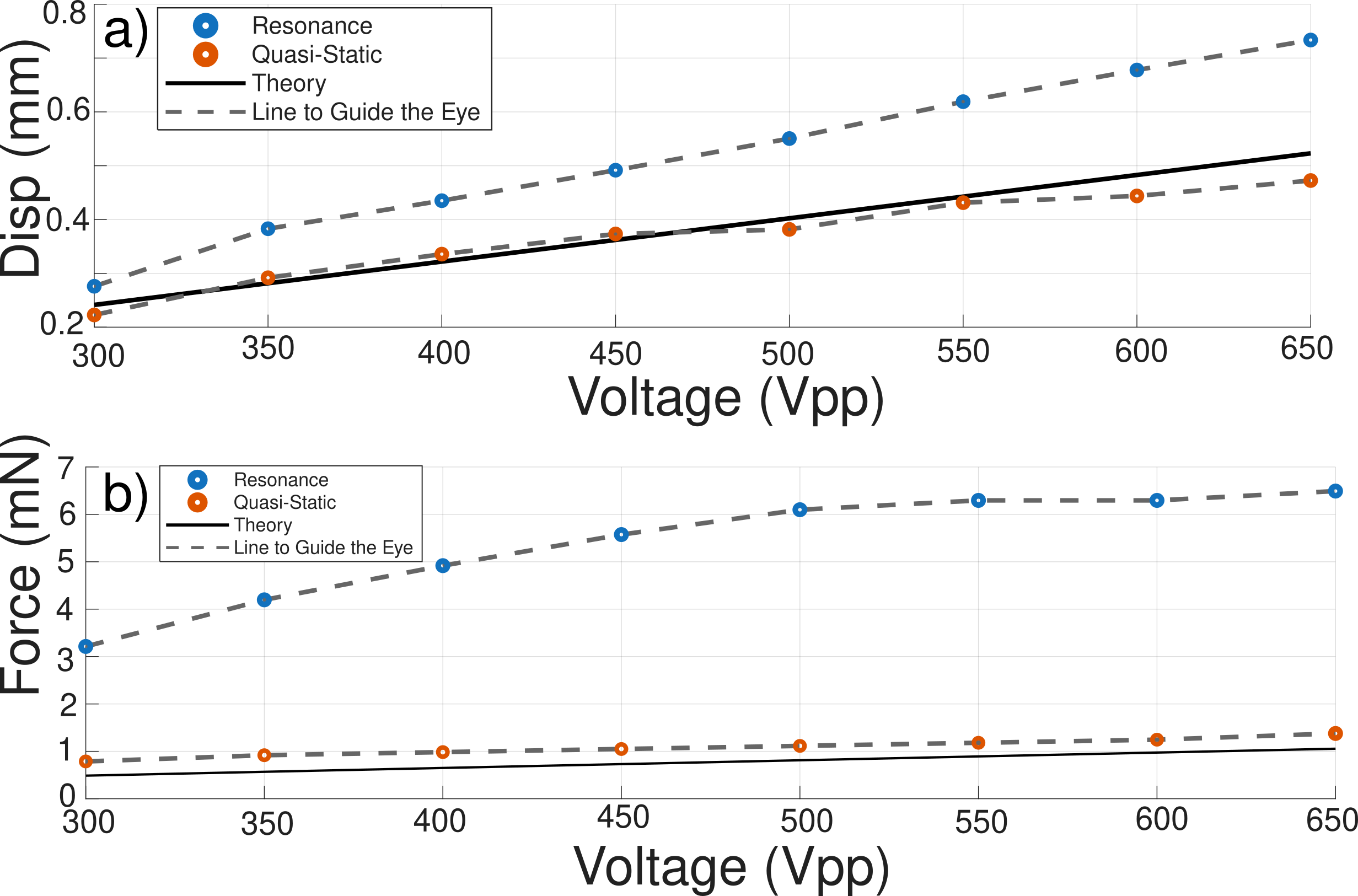}
	\caption{(a) Free displacement and (b) blocked force as a function of signal amplitude and drive frequency.}
	\label{fig:Voltage}
\end{figure}

The performance of the actuator with 4 \SI{12}{\micro\meter} layers was quantified (Fig.~\ref{fig:Voltage}) and compared against the theory derived in Sec.~\ref{sec: Theory}. A \SI{5}{\hertz} sinusoidal signal served as a proxy for quasi-static operation. Each actuator's resonant frequency was determined using a 60-second chirp (\SIrange{1}{600}{\hertz}). 

\begin{figure}[!htb]
	\centering
	\includegraphics[width=0.85\linewidth]{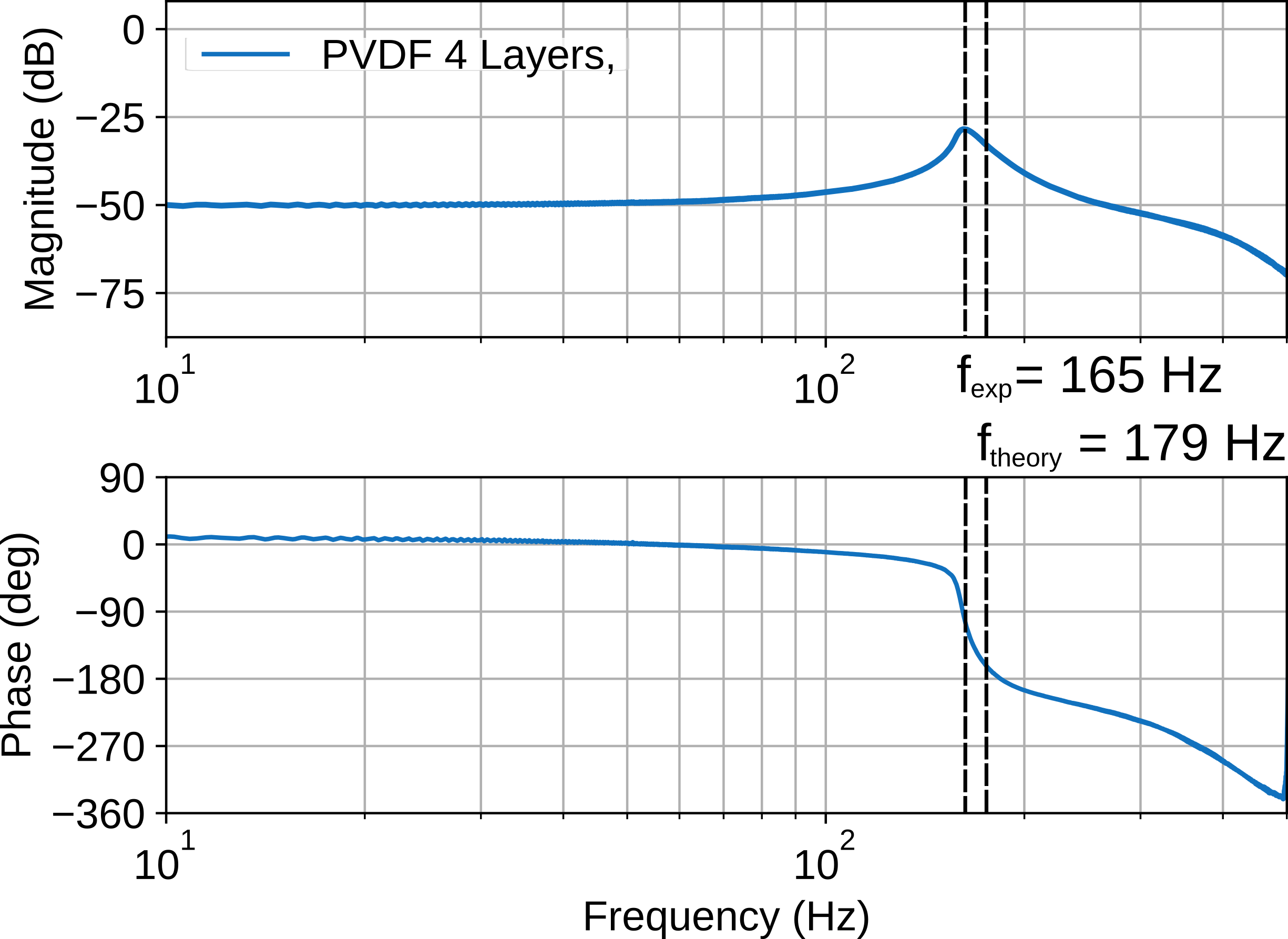}
	\caption{Empirical Bode plot showing resonant amplification of an actuator with 4 \SI{12}{\micro\meter} layers.}
	\label{fig:BodePlot}
\end{figure}

Quasi-static and resonant performance data agreed closely with theoretical predictions (Figs.~\ref{fig:Voltage} and \ref{fig:BodePlot}). Resonant actuation markedly increased both force and displacement at the cost of operating bandwidth (Q $\sim5.73$).
These results suggest the feasibility of actuator customization using first principles (Eq.~\ref{eq:freq}) to match applications \cite{kabutz_mclari_2023, chen_controlled_2019, hartmann2025highly}. 

\subsection{Power density enhancement through layering}
\label{sec:layers}

\begin{figure}[!b]
	\centering
	\includegraphics[width=0.9\linewidth]{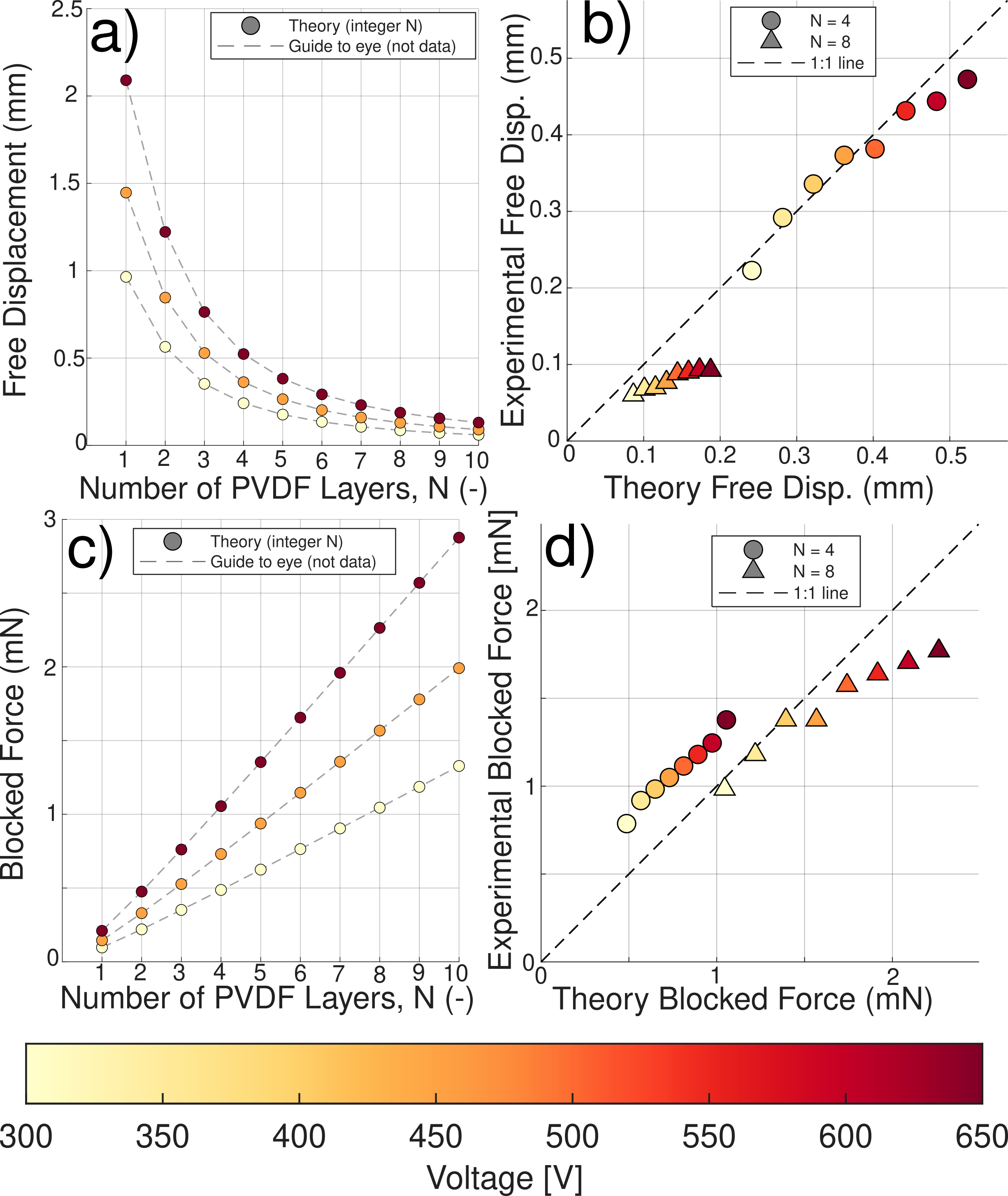}
	\caption{Theoretical predictions of the effect of layer number on (a) free displacement and (c) blocked force. Parity plots show strong agreement between model and experiment for (b) free displacement and (d) blocked force.}
	\label{fig:Theory}
\end{figure}

Under quasi-static conditions and at a given voltage, increasing the number of layers $N$ decreases free displacement ($\delta_{\text{free}} \propto N^{-2}$ for our stiffness regime, Fig.~\ref{fig:Theory}a) and increases blocked force ($F_{\text{b}} \propto N$, Fig.~\ref{fig:Theory}c). These model predictions (Sec.~\ref{sec: Theory}) agreed closely with experiments (Fig.~\ref{fig:Theory}b,d).

\begin{figure}[!htb]
	\centering
	\includegraphics[width=0.95\linewidth]{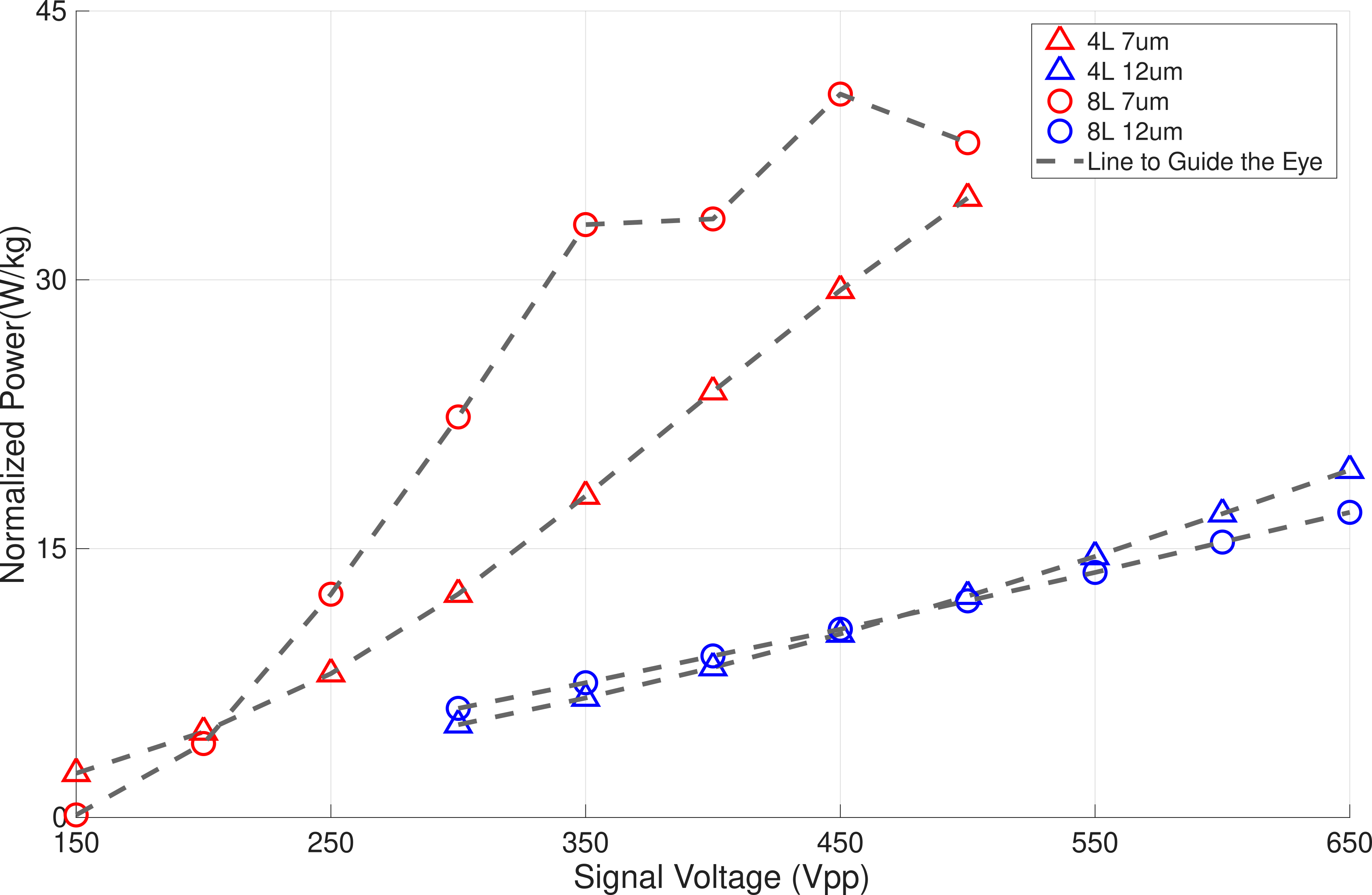}
    \caption{Actuator continuous power density at resonance. Thinner layers produce higher output at the same voltage.}
	\label{fig:powerdensity}
\end{figure}

We also expect similar performance enhancements at resonance and 
measured Q factors ranged from 5.73 to 10.68 across layer combinations.
We observed that thinner layers demonstrate higher continuous power density (Eq.~\ref{eq: power}; Fig.~\ref{fig:powerdensity}) at a lower corresponding voltage owing to reduced mass. 

\subsection{Actuator Comparison}
\label{sec:peakpowerdensity}

We contextualize our actuator performance amongst other state-of-the-art (SoA) millimeter-scale soft actuators in Table \ref{tab:actuator_comparison}. We find that PVDF-based actuators occupy a middle ground between DEA (soft and robust with modest power density) and PZT (fragile and rigid with high power density) and thus is favourable for mobile power and weight limited robotic applications. Excitingly, our actuator performs (voltage-normalized power density, W/kg/V) similarly to SoA PVDF actuators \cite{ZhuTakeoff2026} with much lower voltage requirements. Although current absolute power density (W/kg) is lower than SoA PZT \cite{steltz2008redesign} and DEA \cite{chen_controlled_2019}, recent results \cite{ZhuTakeoff2026} indicate the promise of this approach with further optimization.

\begin{table}[!htb]
    \centering
    \caption{Comparison of electrically driven soft actuator technologies. Values labelled by (*) are not directly reported in cited text, but calculated from the values that are reported.}
    \label{tab:actuator_comparison}
    \footnotesize
    \setlength{\tabcolsep}{3pt}
    \resizebox{\columnwidth}{!}{%
    \begin{tabular}{lccccc}
        \hline
        \textbf{Peak Value} & \textbf{Ours} & \textbf{PVDF SoA}~\cite{ZhuTakeoff2026} & \textbf{PZT H5}~\cite{steltz2008redesign} & \textbf{DEA}~\cite{chen_controlled_2019} & \textbf{Milli-HASEL}~\cite{shmulevich2026microfabrication} \\
        \hline
        (mg) & \num{15} & \num{70} & \num{12} & \num{110} & \num{5} \\
        (V) & 500 & 1400 & \num{200} & 2000 & 300 \\
        (Hz) & 268 & 110 & 275 & 275 & 125 \\
        (mm) & \num{9.2} & 0.65 & 0.52 & 2 & 0.29* \\
        (mN) & \num{20} & 47 & \num{123} & 440 & 0.076* \\
        (W/kg) & \num{633} & 1600 & 819 & 1200 & 0.18* \\
        (W/kg/V) & \num{1.27} & 1.14* & 3.27* & 0.6* & 0.0006* \\
        \hline
    \end{tabular}%
    }
\end{table}

\section{Robotic Demonstrations}
\label{sec: TriBot}

We constructed \textit{PVDF Tripod}, a planar triangular crawling microrobot with three 8-\SI{12}{\micro\meter} actuators were mounted at 120$^\circ$ intervals around a triangular fibreglass (FR4) chassis (\SI{4}{\centi\meter} arm length, \SI{117}{\milli\gram} mass). Each actuator operated as a bending unimorph, directly contacting the ground (without transmissions for motion amplification) to produce locomotion simplifying robot design. 

\begin{figure}[!htb]
    \centering
    \includegraphics[width=0.9\linewidth]{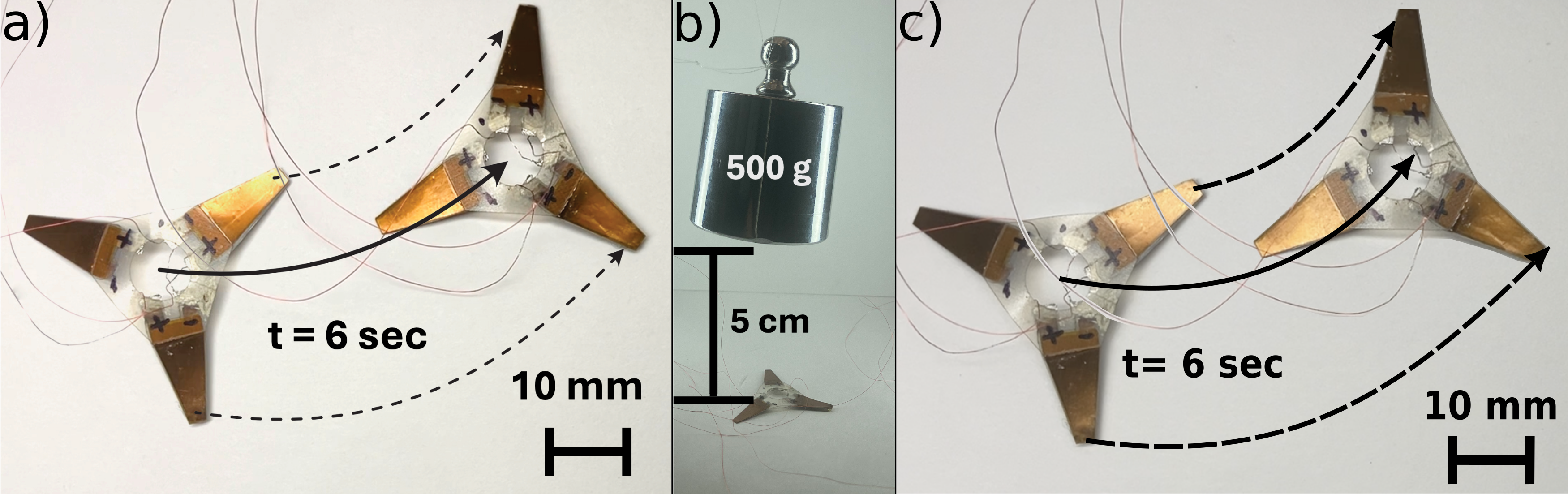}
    \caption{PVDF Tripod translation experiment with weight drop - (a) before, (b) weight on, and c) after. }
    \label{fig:TriBot}
\end{figure}

PVDF Tripod was actuated by sinusoidal voltage waveforms delivered in-phase to two of the three actuators. At low-frequency drive ($<100$ Hz), the robot achieved stepwise hopping, while at 
at resonance (\SI{\sim 200}{\hertz}), demonstrated continuous locomotion achieving top forward speeds of \SI{\sim 6}{\milli\meter/\second} ($\sim$0.15 body lengths/s).
PVDF’s mechanical toughness and compliance provided robustness  \cite{ZhuTakeoff2026}: the actuators sustained repeated ground collisions at resonance without fracture, unlike brittle PZT actuators \cite{chen2026call}, while the robot maintained locomotion capability even after being stepped on (Fig. \ref{fig:RobotOverview}) or crushed by a \SI{500}{\gram} weight (Fig. \ref{fig:TriBot}).

\section{DISCUSSION \& FUTURE WORK}

Our work demonstrates a scalable multilayer PVDF actuation approach with tunable design and out-of-cleanroom fabrication for easy and widespread adoption within the microrobot community. Our actuators perform similar to SoA PVDF actuators and improve on limitations of brittle piezoceramic actuators with a voltage requirement much lower than other PVDF actuators but still higher than PZT.
Table~\ref{tab:actuator_comparison} summarizes representative performance metrics from comparable small-scale actuator technologies. 

Despite the promising initial results, there are several areas of improvement. The use of dedicated adhesive layer limits actuation performance due to inactive material in the layup and also introduced shear lag compared to a more specialized process  without interfacial layers \cite{toinet2024design}. Eliminating the need for adhesive while maintaining a simplified manufacturing method would greatly benefit future iterations of multilaminate actuators. Compliant materials such as gold mesh, silver nanowires, PEDOT:PSS, graphene inks, etc. could extend actuator lifetime under cyclic loading and reduce manufacturing complexity. Conductive ink printing may be a strong choice for batch processing of actuator layers to drive down cost\cite{Cummins2012_Inkjet}. 
Additionally, PVDF's direct electromechanical coupling makes it an ideal material to include sensors that close the loop directly in actuation components \cite{kabutz_integrated_2025}.
Looking forward, multilayer PVDF actuators can be extended to create more capable microrobots including insect-inspired flexural leg systems, 
aerial morphing through wing camber and twist modulation,
amphibious robots that take advantage of the lightweight, haptic interfaces and skins that prioritize conformability for wearables, etc.
These examples highlight PVDF’s potential to expand the actuation toolbox for physical intelligence and distributed actuation in microrobotic systems.





\medskip

\bibliographystyle{ieeetr}
\bibliography{pvdf_actuators.bib} 

\end{document}